\pdfoutput=1
\documentclass[11pt]{article}

\usepackage{emnlp2022}

\usepackage{times}
\usepackage{latexsym}
\usepackage{graphicx}
\usepackage{todonotes}
\usepackage{multirow}
\usepackage{amsfonts}
\usepackage{adjustbox}
\usepackage{amsmath}
\usepackage{ntheorem}
\usepackage{subcaption}

\usepackage{booktabs} % To thicken table lines

\usepackage[T1]{fontenc}
\usepackage[utf8]{inputenc}

\usepackage{microtype}

\title{Joint Multilingual Knowledge Graph Completion and Alignment}

\author{Vinh Tong$^1$, Dat Quoc Nguyen$^2$, Trung Thanh Huynh$^3$, Tam Thanh Nguyen$^4$, \\ \textbf{Quoc Viet Hung Nguyen$^4$, Mathias Niepert$^1$}\\
  $^1$University of Stuttgart, Germany; $^2$VinAI Research, Vietnam; \\ $^3$EPFL, Switzerland;  $^4$Griffith University, Australia\\
  $^1$ \texttt{vinh.tong@ipvs.uni-stuttgart.de}, $^2$\texttt{v.datnq9@vinai.io} }

\begin{document}

\maketitle

\begin{abstract}

Knowledge graph (KG) alignment and completion are usually treated as two independent tasks. While recent work has leveraged entity and relation alignments from multiple KGs, such as alignments between multilingual KGs with common entities and relations, a deeper understanding of the ways in which multilingual KG completion (MKGC) can aid the creation of multilingual KG alignments (MKGA) is still limited. Motivated by the observation that structural inconsistencies -- the main challenge for MKGA models -- can be mitigated through KG completion methods, we propose a novel model for jointly completing and aligning knowledge graphs. The proposed model combines two components that jointly accomplish KG completion and alignment. These two components employ relation-aware graph neural networks that we propose to encode multi-hop neighborhood structures into entity and relation representations. Moreover, we also propose (i) a structural inconsistency reduction mechanism to incorporate information from the completion into the alignment component, and (ii) an alignment seed enlargement and triple transferring mechanism to enlarge alignment seeds and transfer triples during KGs alignment. Extensive experiments on a public multilingual benchmark show that our proposed model outperforms existing competitive baselines, obtaining new state-of-the-art results on both MKGC and MKGA tasks.

\end{abstract}

\section{Introduction}

Knowledge graphs (KGs) represent facts about real-world entities as triples of the form $(\mathtt{head\_entity}, \mathtt{relation\_type}, \mathtt{tail\_entity})$. KGs are widely used in numerous applications and research domains such as dialogue systems~\cite{jung2020attnio}, machine translation \cite{zhao2020knowledge}, and electronic medical records~\cite{rotmensch2017learning}. Almost all KGs, however, even those at the scale of billions of triples, are far from being complete. This motivates KG completion (KGC) which aims to derive missing triples from an incomplete knowledge graph~\cite{bordes2013translating, wang2014knowledge,yang2014embedding, pmlr-v48-trouillon16, liu2017analogical,dettmers2018convolutional,Nguyen2020KGC,Ji_2021,tong2021two}.

\begin{figure*}[!t]
	\centering
	\includegraphics[width=0.75\linewidth]{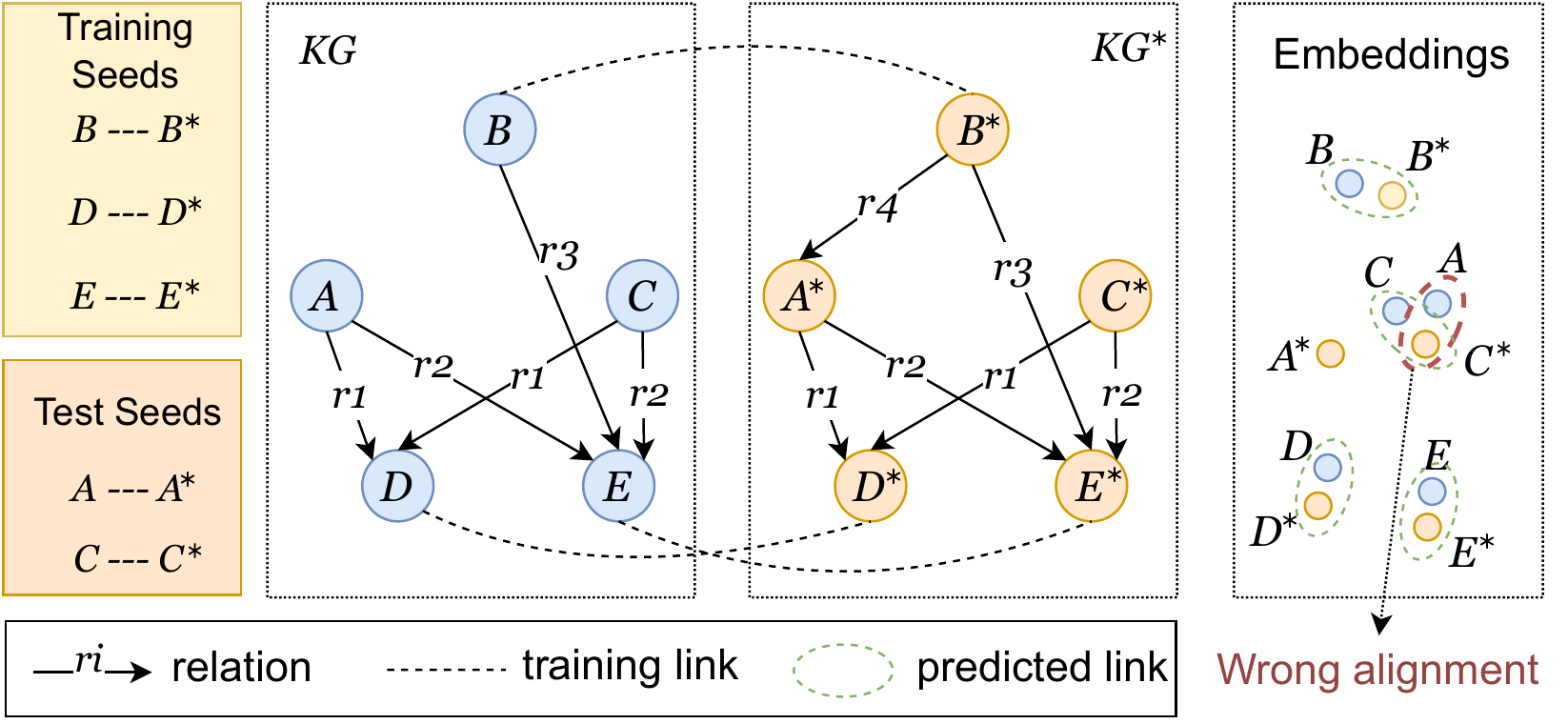}
    \caption{The incompleteness of $KG$ (missing triple $(B, r4, A)$) might lead to a wrong alignment prediction (i.e. both $A$ and $C$ are predicted to be aligned to $C^*$). If $A$ and $A^*$ were aligned, however, the missing triple $(B, r4, A)$ in $KG$ could be found by transferring triple $(B^*, r4, A^*)$ from $KG^*$.}
    \label{fig:Example}
    % \vspace{0em}
 \end{figure*}

Most popular KGs such as YAGO \cite{Suchanek:2007}, BabelNet \cite{navigli-ponzetto-2010-babelnet}, and DBpedia \cite{LehmannIJJKMHMK15}, are multilingual, that is, they contain sets of triples constructed from sources in different languages. Fortunately, these KGs often complement each other since a KG in one language might be more comprehensive in some domains compared to a KG in a different language, and vice versa, while still sharing a large number of entities and relation types \cite{sun2020dual}. Especially KGs for low-resource languages could benefit from triples contained in KGs for high-resource languages. Although numerous approaches to KGC have been proposed in recent years \cite{bordes2013translating, dettmers2018convolutional, vashishth2019composition}, most of these only operate on one KG at a time. Treating KGs independently, however, might lead to poor performance due to the sparseness of low-resource languages. Motivated by this observation, some methods have tried to improve multilingual KG completion (MKGC) using multilingual KG alignment (MKGA) \cite{chen2020multilingual, singh2021multilingual, huang2022multilingual}.

The alignment problem is challenging due to the varying levels of completeness of monolingual KGs \cite{sun2020benchmarking}. The resulting structural inconsistency between KGs leads to the problem of corresponding entities in two KGs having vastly different embeddings \citep{jumping_knowledge}. 
\autoref{fig:Example} illustrates that, in principle, MKGC and MKGA should be mutually beneficial. Given the alignment seed set $\{(E, E^*), (D, D^*), (B, B^*) \}$, the task here is to find corresponding entities for $A$ and $C$ which are $A^*$ and $C^*$, respectively. The two KGs share a similar structure but the triple $( B^*, r4, A^*)$ in $KG^*$  has no corresponding triple in $KG$. This causes the embeddings of entities $A$ and $A^*$ to differ and, thus, makes it difficult to identify the alignment between $A$ and $A^*$. Indeed, $A$ is more likely to be aligned to $C^*$ as they share a comparable local structure (similar degree and 1-hop neighbor set). Thus, completing missing triples is crucial to improve the alignment quality. Indeed, if one aligned $A$ to $A^*$, one could recover $(B, r4, A)$ by transferring  $(B^*, r4, A^*)$ from $KG^*$.

Motivated by these observations, we propose \textbf{JMAC}, a method for \textbf{J}oint \textbf{M}ultilingual KG \textbf{C}ompletion and \textbf{A}lignment, consisting of two interdependent Completion and Alignment components. 
Both components employ relation-aware graph neural networks (GNNs) to encode multi-hop neighborhood information into entity and relation embeddings. The Completion component is trained to reconstruct missing triples using the TransE translation-based loss \cite{bordes2013translating} and an additional loss term that incorporates information about the already known alignments. While we learn separate  embeddings for the Alignment and Completion components, the embeddings of the Completion component are used within the Alignment component, mitigating the aforementioned problem of structural inconsistencies. In addition, we propose a mechanism for estimating the alignment entropy which is used to adaptively and iteratively grow the alignment seed set. Finally, we also propose a method for transferring triples based on the currently derived alignments.

Our contributions are as follows: % (i) We propose JMAC, a two-component architecture consisting of Completion and Alignment components for joint multilingual KG completion and alignment; (ii) we propose a relation-aware GNN for KG embeddings which learns representations for alignment and completion tasks; (iii) we introduce a structural inconsistency reduction mechanism that fuses embeddings from the Completion component with those of the Alignment component; (iv) we propose an alignment seed enlargement and triple transfer mechanism; and (v) we conduct extensive experiments using  the public multilingual benchmark DBP-5L \cite{chen2020multilingual} and show that our model outperforms existing competitive baselines and achieves  state-of-the-art results on both MKGC and MKGA tasks.

\begin{itemize}
\item We propose JMAC, a two-component architecture consisting of Completion and Alignment components for joint multilingual KG completion and alignment.

\item We propose a relation-aware GNN for KG embeddings, which learns representations for alignment and completion tasks. 

\item We introduce a structural inconsistency reduction mechanism that fuses embeddings from the Completion component with those of the Alignment component.

\item We propose an alignment seed enlargement and triple transfer mechanism.

\item We conduct extensive experiments using  the public multilingual benchmark DBP-5L \cite{chen2020multilingual} and show that our model outperforms existing competitive baselines and achieves  state-of-the-art results on both MKGC and MKGA tasks.
\end{itemize}

\begin{figure*}[!t]
	\centering
	\includegraphics[width=0.75\linewidth]{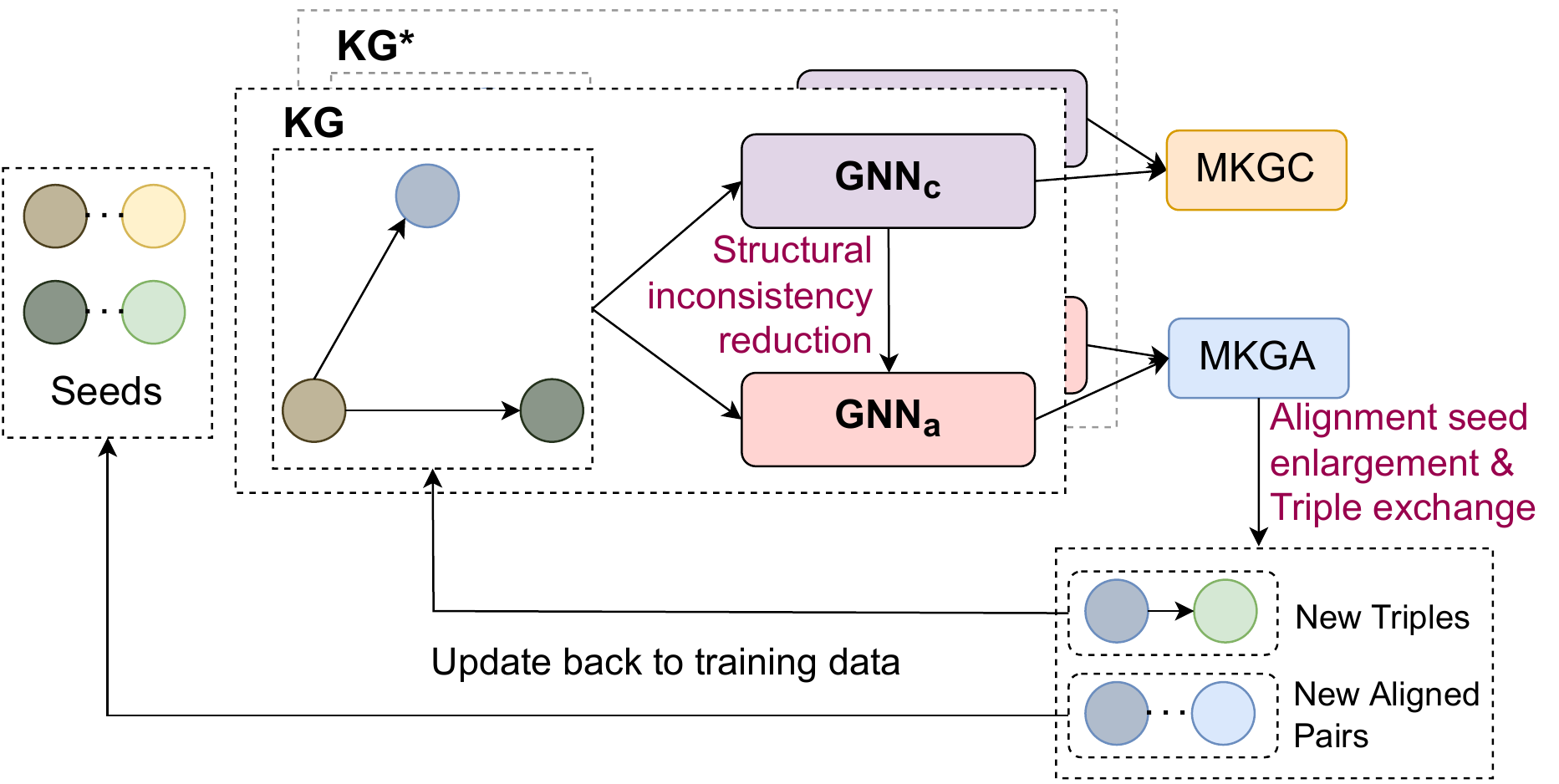}
    \caption{The overall architecture of JMAC}
    \label{fig:framework}
 \end{figure*}

\section{Problem Definition and Related Work}

Let  $\mathcal{G} = (\mathcal{E}, \mathcal{R}, \mathcal{T})$ denote a KG, where $\mathcal{E}$,  $\mathcal{R}$ and $\mathcal{T}$ denote the sets of entities, relations, and triples, respectively. A triple $(e_h, r, e_t) \in \mathcal{T}$ is an atomic unit, which represents some relation $r \in \mathcal{R}$ between a head entity $e_h \in \mathcal{E}$ and a tail entity $e_t \in \mathcal{E}$. 

\subsection{Multilingual KG completion (MKGC)}\label{ssec:mkgc}

Given a KG $\mathcal{G} = ({\mathcal{E}}, {\mathcal{R}}, {\mathcal{T}})$, the KG completion (KGC) task aims to predict missing triples $(e_h, r, e_t)$, that is, to predict the missing tail entity $e_t \in \mathcal{E}$ of an incomplete triple $(e_h, r, ?)$ or the missing head entity $e_h \in \mathcal{E}$ of an incomplete triple $(?, r, e_t)$, where ? denotes the missing element. 

Embedding models for KG completion have been proven to give state-of-the-art results, representing entities and relation types with latent feature vectors, matrices, and/or third-order tensors \cite{Nguyen2020KGC,Ji_2021}. These models define a score function $f$ and are trained to make the score $f(e_h, r, e_t)$ of a correct triple $(e_h, r, e_t)$ larger than the score $f(e_{h'}, r', e_{t'})$ of an incorrect or not known to be correct triple $(e_{h'}, r', e_{t'})$. 
The earliest instances of these embedding models use shallow neural networks with translation-based score functions \cite{bordes2013translating, wang2014knowledge, lin2015learning}. Recently, KGC approaches using deep embedding models and more complex scoring functions have been proposed, such as CNN-based models \cite{dettmers2018convolutional, nguyen2017novel}, RNN-based models \cite{liu2017analogical, guo2018dskg}, and GNN-based models \cite{SchlichtkrullKB17,shang2019end, vashishth2019composition,nguyen2021node}.

The MKGC task is to perform the KGC task on a KG given the availability of other KGs in different languages \cite{chen2020multilingual,huang2022multilingual}.

\subsection{Multilingual KG alignment (MKGA)}\label{ssec:mkga}

MKGA, which is also known as cross-lingual entity alignment, aims to match entities with their counterparts from KGs in different languages \cite{MTransE,wang2018cross,wu2019relation, sun2020knowledge}. Without loss of generalization, we define the alignment between a source graph $\mathcal{G} = ({\mathcal{E}}, {\mathcal{R}}, {\mathcal{T}})$ and a target graph $\mathcal{G}^* = ({\mathcal{E}}^*, {\mathcal{R}}^*, {\mathcal{T}}^*)$. For each entity $e \in \mathcal{E}$, the MKGA task is now to find $e^* \in \mathcal{E}^*$ (if any). 
%The matched entity pairs $(e, e^*)$ are referred to as anchor links.  

Existing models compute an alignment matrix whose elements represent the similarity score between any two entities  $e \in \mathcal{E}$ and $e^* \in \mathcal{E}^*$ across two KGs. The models then employ a greedy matching algorithm \cite{greedymatch} to infer matching entities from the alignment matrix. The models typically require an alignment seed set $\mathbb{L}$ of pre-aligned entity pairs $(e, e^*)$.

\subsection{Joint MKGC and MKGA} 

The joint MKGC and MKGA problem aims to infer both, new triples for each KG and new aligned entity pairs for each pair of KGs. Performing two tasks jointly might be beneficial:  missing triples $(e_h, r, e_t) $ in one KG could be recovered by cross-checking another KG via the alignment, which, in turn, could be boosted by the newly added triples. 

Despite the obvious benefit to complete and align KGs jointly, there has not been much work addressing the problem. A notable exception is the application of multi-task learning to the problem \cite{singh2021multilingual}.  The proposed multi-task model, however, is not able to capture local neighborhood information. Another limitation is the  missing robustness to the previously described issue of structural inconsistencies between the  KGs during training.

\section{JMAC: Joint Multilingual Alignment and Completion}

\autoref{fig:framework} illustrates the architecture of \textbf{JMAC} which consists of the Completion and Alignment components, respectively. Each component uses a relation-aware GNN to encode multi-hop neighborhood information into entity and relation embeddings. The use of two GNN encoders is beneficial as embeddings that are suitable for the alignment might differ from those that are most beneficial for the completion problem.

\subsection{Relation-aware graph neural network}\label{ssec:relationgnn}

To better capture relation information, we propose a GNN architecture that uses relation-aware messages and relation-aware attention scores. 

We unify heterogeneous information from KGs using a GNN with $K$ layers. For the $k$-th layer (denoted by the superscript $^k$), we update the representation $\mathbf{a}_e^{k + 1}$ of each entity $e \in \mathcal{E}$ as:

\setlength{\abovedisplayskip}{0pt}
\setlength{\belowdisplayskip}{5pt}
\begin{equation}
    \mathbf{a}_e^{k + 1} = g \left( \sum_{(e', r) \in \mathcal{N}(e)} \alpha_{e,e',r}^k \mathbf{m}^k_{e',r} + \mathbf{a}_e^{k} \right)
    \label{eqn:entity_update}
\end{equation}

\noindent where $\mathbf{a}_e^{k} \in \mathbb{R}^n$ is the vector representation of entity $e$ at the $k$-th layer; $\mathcal{N}(e) = \left\{(e', r) | ( e, r, e' ) \in \mathcal{T} \cup ( e', r, e ) \in \mathcal{T} \right\}$ is the neighbor set of entity $e$; and the vector $\mathbf{m}^k_{e',r} \in \mathbb{R}^n$ denotes the message passed from neighbor entity $e'$ to entity $e$ through relation $r$. Here,  $\alpha^k_{e,e',r}$ represents the attention weight that regulates the importance of the message $\mathbf{m}^k_{e',r}$ for entity $e$; and $g(.)$ is a linear transformation followed by a $\mathsf{Tanh}$ function. 

The innovations of our GNN-based model, which we describe in more detail in the following two sections, are (i) we make the message-passing NN to be relation-aware by learning the relation embedding $\mathbf{a}^{k}_r \in \mathbb{R}^n$ for each relation $r \in \mathcal{R}$ and by integrating it into the entity message passing scheme $\mathbf{m}^k_{e',r}$ and (ii) we introduce an attention weight $\alpha^k_{e,e',r}$ to further enhance the relation-aware capability of our GNN-based embeddings.

\paragraph{Relation-aware message} Unlike existing GNN-based approaches that  infer relation embeddings from learned entity embeddings \cite{sun2020knowledge}, our approach allows entity and relation embeddings to be learned jointly and, thus, to both contribute to the message passing neural network. This is achieved through an entity-relation composition operation. The message $\mathbf{m}^k_{e',r}$ in Equation \ref{eqn:entity_update} is defined  as:

\begin{align}
\label{eqn:ramessage}
    \mathbf{m}^k_{e',r} = \mathbf{a}^k_{e'} - \mathsf{MLP}^k_{\text{comp}}\left(\mathbf{a}^k_r \right) 
\end{align}

\noindent where $\mathsf{MLP}^k_{\text{comp}}: \mathbb{R}^n \rightarrow \mathbb{R}^n$ is a two-layer MLP with the $\mathsf{LeakyReLU}$ activation function.  

Here, the relation embedding is updated by

\begin{equation}
    \mathbf{a}_r^{k + 1} = \mathsf{MLP}^k_{\text{rel}} \left( \mathbf{a}_r^k \right)
    \label{eqn:relation_update}
\end{equation}

\noindent where $\mathsf{MLP}^k_{\text{rel}}: \mathbb{R} ^ n \rightarrow \mathbb{R} ^ n$ maps relations to a new embedding space and allows them to be utilized in the next layer.

\paragraph{Relation-aware attention} We define the weight $\alpha^k_{e,e',r}$ in \autoref{eqn:entity_update} to be relation-aware as:

{\small
\begin{equation}
\label{eqn:ra_attention}
    \alpha_{e, e', r}^k = \frac{\exp\left(\mathsf{MLP}_{\text{att}}^k\left(\mathbf{a}_e^k \circ \mathbf{m}^k_{e', r}\right)\right)
    }{\sum\limits_{(e", r") \in \mathcal{N}(e)} \exp\left(\mathsf{MLP}_{\text{att}}^k\left(\mathbf{a}_e^k \circ \mathbf{m}^k_{e", r"}\right)\right)} 
\end{equation}
}

\noindent where $\mathsf{MLP}_{\text{att}}^k: \mathbb{R}^{2\times n} \rightarrow \mathbb{R}$;  and $\circ$ denotes the vector concatenation operator. As the message vector $\mathbf{m}_{e',r}^k$ contains the information of not only neighbor entity $e'$ but also neighbor relation $r$, our attentive score $\alpha_{e,e',r}^k$ can capture the importance of the message coming from entity $e'$ to entity $e$ conditioned on relation $r$ connecting them.

\paragraph{Notation extension} Recall that we use two different relation-aware GNN encoders as illustrated in Figure \ref{fig:framework}: one for the Completion and one for the Alignment component. To distinguish these two encoders, we use  $\mathbf{c}$ and $\mathbf{a}$ to denote the embedding representations used for the Completion and Alignment components, respectively. In particular, $\mathbf{a}_e^{k}$ and $\mathbf{a}_r^{k}$ are now the corresponding embeddings of entity $e$ and relation $r$ at the $k$-th layer of the relation-aware GNN in the Alignment component. Furthermore, $\mathbf{c}_e^{k}$ and $\mathbf{c}_r^{k}$ are the corresponding embeddings of entity $e$ and relation $r$ at the $k$-th layer of the relation-aware GNN in the Completion component (computed as those of the Alignment component defined in equations \ref{eqn:entity_update} and \ref{eqn:relation_update}). 

\subsection{Completion component}\label{ssec:completion}

The Completion component  works similarly to KG embedding models \cite{Nguyen2020KGC,Ji_2021} which compute a score $f(e_h, r, e_t)$ for each triple $(e_h, r, e_t)$. Our score function $f$ is based on  TransE \cite{bordes2013translating} and is computed across all hidden layers of the relation-aware GNN encoder in the Completion component as follows:

\begin{align}
f^k(e_h, r, e_t) &= -\|\mathbf{c}^k_{e_h} + \mathbf{c}^k_r - \mathbf{c}^k_{e_t}\|_{\ell_{1}} \\
 f(e_h, r, e_t) &= \sum_k f^k(e_h, r, e_t) \label{equa:scorefunction}
\end{align}

We use a margin-based pairwise ranking loss \cite{bordes2013translating} across all hidden layers:

{\small
\begin{equation}
    % \nonumber
    \mathcal{L}_{c\_1} = \sum_{ \substack{k \\ (e_h, r, e_t) \in \mathcal{T} \\ (\overline{e}_h, r,  \overline{e}_t) \in \overline{\mathcal{T}}}} {[\gamma_c - f^k(e_h, r, e_t)} +
    f^k(\overline{e}_h, r,  \overline{e}_t)]_+ 
    \label{eqn:complossterm1}
\end{equation}
}

\noindent where $[x]_+=\mathsf{max}(0, x)$; $\gamma_c > 0$ is the margin hyper-parameter; and $\overline{\mathcal{T}}$ is the set of incorrect triples constructed by corrupting either the head or the tail entity the correct triple $(e_h, r, e_t) \in \mathcal{T}$.

Also, given the availability of the alignment seed set $\mathbb{L}$ of pre-aligned entity pairs $(e, e^*)$ among KGs as mentioned in Section \ref{ssec:mkga}, we additionally compute the following alignment constraint loss:

\begin{equation}
    % \nonumber
    \mathcal{L}_{c\_2} =  \sum_{k} \sum_{(e, e^*)\in \mathbb{L}}  \mathsf{d}_{\mathtt{cos}}\left(\mathbf{c}^k_e, \mathbf{c}^k_{e^*}\right)
    \label{eqn:complossterm2}
\end{equation}

\noindent where $\mathsf{d}_{\mathtt{cos}}$ denotes the cosine distance. 

To incorporate alignment information into the Completion component, our MKGC loss is computed as the sum of the two losses $\mathcal{L}_{c\_1}$ and  $\mathcal{L}_{c\_2}$: 

\begin{equation}
    \mathcal{L}_{c} = \mathcal{L}_{c\_1} + \mathcal{L}_{c\_2}
\end{equation}

\subsection{Alignment component}\label{ssec:alignment}

The Alignment component is to perform the MKGA task as defined in Section \ref{ssec:mkga}. 
We propose to incorporate a structural inconsistency reduction inherited from the Completion into the Alignment component. We also introduce a mechanism to enlarge the alignment seed set $\mathbb{L}$.

\paragraph{Structural inconsistency reduction (SIR)} The different completeness levels of KGs lead to structural inconsistencies that might cause incorrect alignment predictions. Fortunately, entity and relation embeddings  in the Completion component (i.e. $\mathbf{c}_e^{k}$ and $\mathbf{c}_r^{k}$) could help reconstruct missing triples,  reducing the structural inconsistencies between KGs. To this end, we propose to  incorporate the Completion component embeddings at the $k$-th layer of the relation-aware GNN in the Alignment component as follows:

\begin{align}
    & {{{{\mathbf{a}^k_e}}}} = {{\mathsf{MLP}}}^k_{\text{a\_1}}\left({\mathbf{c}^k_e} \circ {\mathbf{a}^k_e} \right)\ \ \ \ \forall e \label{eqn:aesir}\\
    & {{{\mathbf{a}}}^k_r} = {{\mathsf{MLP}}}^k_{\text{a\_2}} \left( {\mathbf{c}^k_r}  \circ {\mathbf{a}^k_r} \right)\ \ \ \ \forall r \label{eqn:arsir}
\end{align}

\noindent where $\mathsf{MLP}_{\text{a\_1}}^k: \mathbb{R}^{2\times n} \rightarrow \mathbb{R}^n$; and $\mathsf{MLP}_{\text{a\_2}}^k: \mathbb{R}^{2\times n} \rightarrow \mathbb{R}^n$. The transformed embeddings then will be used as input for the next layer following equations \ref{eqn:entity_update} and \ref{eqn:relation_update}. 
This allows the structural inconsistency reduction to take place at every GNN layer of the two components, enabling their deep integration. %Note that an analogous mechanism could be applied to incorporate the Alignment component embeddings into the Completion component. However, we do not perform such incorporation because the Alignment component embeddings might bring unnecessary (noise) information to the Completion component.

\paragraph{Final  entity and relation embeddings for MKGA} We compute the final embeddings for entities and relations for this Alignment component as follows:

\begin{align}
    \mathbf{a}_e = \mathsf{MLP}_{\text{a\_3}} \left( \mathbf{a}_e^0 \circ \mathbf{a}_e^1 \circ ... \circ \mathbf{a}_e^K \right) \\
    \mathbf{a}_r = \mathsf{MLP}_{\text{a\_4}} \left(  \mathbf{a}_r^0 \circ \mathbf{a}_r^1 \circ ... \circ \mathbf{a}_r^K \right)
    \label{eqn:all_rel}
\end{align}

\noindent where $\mathsf{MLP}_{\text{a\_3}}: \mathbb{R}^{(K+1)\times n} \rightarrow \mathbb{R}^n$; and $\mathsf{MLP}_{\text{a\_4}}: \mathbb{R}^{(K+1)\times n} \rightarrow \mathbb{R}^n$.

%%%%%%%%%%%%%%%%%%
\begin{table*}[!t]
%\begin{adjustbox}{width=0.975\linewidth,center}
    \centering
    \def\arraystretch{1.1}
    \begin{tabular}{l|l|l|l|l|l}
    \hline
         \bf Language & \bf Greek & \bf Japanese & \bf French & \bf Spanish & \bf English \\
         \hline
         \bf \#Relation & 111 & 128 & 144 & 178 & 831 \\
         \hline
         \bf \#Entity & 5,231 & 11,805 & 12,382 & 13,176 & 13,996 \\
         \hline
         \bf \#Triple & 13,839 & 28,744 & 54,066 & 49,015 & 80,167 \\
         \hline 
         %%\bottomrule
    \end{tabular}
%    \end{adjustbox}
    \caption{Statistics of DBP-5L.}
    \label{tab:datasetstats}
\end{table*}
%%%%%%%%%%%%%%%%

\paragraph{Alignment seed \underline{en}largement and Triple \underline{tr}ansferring (EnTr)} As mentioned in Section \ref{ssec:mkga}, existing alignment models require an alignment seed set $\mathbb{L}$ of pre-aligned entity pairs for training. An intuitive approach to improve alignment results is to iteratively increase the size of $\mathbb{L}$ during training. The number of alignments by which the size of $\mathbb{L}$ is increased should depend on some notion of certainty the Alignment component has in its predictions. For example, in the early stages of the alignment process, when the KGs are still sparsely connected, $\mathbb{L}$ should be smaller. The more confident the Alignment component is in its predictions, the larger should $\mathbb{L}$ be.  
Hence, we propose EnTr, a method to enlarge the  alignment seed set. EnTr estimates an optimal number of new entity pairs to be added to $\mathbb{L}$ according to a measure of alignment certainty. At each training epoch, EnTr first computes an alignment matrix $\mathbf{A}$, where each element is the cosine similarity between any two entity embeddings of the two KGs, that is, $\mathbf{A}(e, e^*) = 1 - \mathsf{d}_{\mathtt{cos}}({{\mathbf{a}_e}}, {\mathbf{a}_{e^*}})$. EnTr then computes the Shannon entropy of the softmax distribution over this alignment matrix:

\begin{align}
   \mathsf{P}(e^*|e) &= \frac{\exp\left(\mathbf{A}(e, e^*)\right)
    }{\sum_{e^\star \in \mathcal{E}^*} \exp\left(\mathbf{A}(e, e^\star)\right)} \\
   \mathsf{H}(\mathbf{A}) &= -\sum_{e \in \mathcal{E}} \sum_{e^* \in \mathcal{E}^*} \mathsf{P}(e^*|e)\ \mathsf{log}\ \mathsf{P}(e^*|e)
\end{align}

Since the entropy measures the uncertainty of the alignment predictions -- lower entropy corresponds to a smaller alignment uncertainty -- EnTr implements an uncertainty-mediated estimation of the size $\mathsf{q}$ of $\mathbb{L}$ as follows: 

\begin{equation}
    \mathsf{q} = \left\lfloor \beta \cdot  \frac{\mathsf{H}(\mathbf{\widetilde A}) - \mathsf{H}(\mathbf{A})}{\mathsf{H}(\mathbf{\widetilde A})} \cdot \mathsf{min}(|\mathcal{E}|, |\mathcal{E}^*|) \right\rfloor
    \label{eqn:num_sample}
\end{equation}

\noindent 
where $\mathbf{\widetilde A}$ is the alignment matrix before training and $\beta \in [0, 1]$ is a hyper-parameter controlling the number of new entity pairs to generate. 
EnTr then chooses the $\mathsf{q}$  entity pairs with the $\mathsf{q}$ highest cosine similarity scores from $\mathbf{A}$ and sets $\mathbb{L}$ to contain these entity pairs. 

EnTr also performs transfer of triples between the KGs that logically follow from the existing alignments. In particular, for each two aligned entity pairs $(e, e^*)$ and $(e', {e'}^*)$, if $r$ is a relation connecting $e$ and $e'$, EnTr connects $e^*$ and ${e'}^*$ by $r$ as well. This allows the two KGs' structures to become gradually more similar over time.

\paragraph{Optimization}
The learning objective is to minimize the distance between correctly aligned entity pairs while maximizing the distance between negative entity pairs using a  margin-based pairwise ranking loss:

\begin{equation}
    \mathcal{L}_{a} = \sum_{\substack{(e, e^*) \in \mathbb{L} \\ (\overline{e}, \overline{e}^*) \in \overline{\mathbb{L}}}} [\gamma_{a} + \mathsf{d}_{\mathtt{cos}}(\mathbf{a}_e, \mathbf{a}_{e^*}) - \mathsf{d}_{\mathtt{cos}}(\mathbf{a}_{\overline{e}}, \mathbf{a}_{\overline{e}^*})]_+
    \label{eqn:aligncomp_loss}
\end{equation}

\noindent where $\gamma_{a} > 0$ is the margin hyper-parameter; and $\overline{\mathbb{L}}$ is the set of negative entity pairs, which is constructed by replacing one entity of each correctly aligned pair by its nearest entities \cite{wu2019relation}.

\subsection{Model training}\label{ssec:training}

We optimize the losses $\mathcal{L}_{c}$ and $\mathcal{L}_{a}$ iteratively, using two optimizers respectively for the Completion and Alignment losses.  In particular, we hold the Alignment component's parameters fixed and optimize only the loss  $\mathcal{L}_{c}$. Then we hold the Completion component's parameters fixed and only optimize the loss $\mathcal{L}_{a}$. We keep iterating this process in each training epoch.

\section{Experimental Setup} %\todo{Section 4 is to be shorten \& move details into Appendix if over-page}

\subsection{Dataset}

Following previous work \cite{singh2021multilingual,huang2022multilingual}, we conduct experiments using the benchmark DBP-5L\footnote{\url{https://github.com/stasl0217/KEnS/tree/main/data}} \cite{chen2020multilingual}, publicly available for both MKGC and MKGA tasks.\footnote{\newcite{huang2022multilingual} create another multilingual benchmark named E-PKG (to be released at \url{https://github.com/amzn/ss-aga-kgc}), however, it is not yet available as of 24/06/2022, EMNLP 2022's submission deadline.}  DBP-5L consists of 1,392 relations, 56,590 entities, and 225,831 triples across the five languages Greek (\textbf{EL}), Japanese (\textbf{JA}), French (\textbf{FR}), Spanish (\textbf{ES}), and English (\textbf{EN}). \autoref{tab:datasetstats} presents statistics for each DBP-5L language.  Here, each language is referred to as a KG. The DBP-5L benchmark is created for MKGC evaluation. However, each language pair also has pre-aligned entity pairs as the alignment seeds. In particular, on average, about 40\% of the entities in each KG have their counterparts at other KGs. Thus, it can also be used for MKGA evaluation \cite{singh2021multilingual}.  
Similar to prior works, we use the same split of training, validation, and test data, for MKGC available in DBP-5L for each KG. For MKGA, we use the same 50-50 split of the alignment seeds for training and test, as used in  AlignKGC \cite{singh2021multilingual}.\footnote{JMAC can perform KGA on a benchmark that is purely constructed for the KGA task. We show state-of-the-art results obtained for JMAC on a KGA benchmark in the Appendix.}

\subsection{Evaluation protocol}

For MKGC, and following previous work \cite{chen2020multilingual, singh2021multilingual,huang2022multilingual}, each correct test triple $(e_h, r, e_t)$ is corrupted by replacing the tail entity $e_t$ with each of the other entities in turn, and then the correct test triple and corrupted ones are ranked in descending order of their score. Similar to the previous work, before ranking, we also applied the ``Filtered'' setting protocol \cite{bordes2013translating}. We employ standard evaluation metrics, including the mean reciprocal rank (\textbf{MRR}), \textbf{Hits@1} (i.e. the proportion of correct test triples that are ranked first) and \textbf{Hits@10} (i.e. the proportion of correct test triples that are ranked in the top 10 predictions). Here, a higher score reflects better prediction result.

For MKGA, and following previous work \cite{MTransE,wu2019relation,sun2020knowledge,singh2021multilingual}, each correct test pair $(e, e^*)$, where $e \in \mathcal{E}$ and $e^* \in \mathcal{E}^*$, is corrupted by replacing entity $e^*$ with each of the other entities from $\mathcal{E}^*$ in turn, and then the correct test pair and corrupted ones are ranked in descending order of
their similarity score. We also employ the evaluation metrics MRR, Hits@1 and Hits@10.

\begin{table*}[!t]
    \begin{adjustbox}{width=440pt,center}
    \centering
    \setlength{\tabcolsep}{0.2em}
    \def\arraystretch{1.1}
    \begin{tabular}{l|l|ccc|ccc|ccc|ccc|ccc}
    %\hline
    \hline
    \multirow{2}{*}{\bf Method} & 
    \multirow{2}{*}{\bf Align.} &
    \multicolumn{3}{c|}{\bf Greek} &
    \multicolumn{3}{c|}{\bf Japanese} &
    \multicolumn{3}{c|}{\bf French} &
    \multicolumn{3}{c|}{\bf Spanish} &
    \multicolumn{3}{c}{\bf English} \\
    \cline{3-17} 
    & & H@1 & H@10 & MRR & H@1 & H@10 & MRR & H@1 & H@10 & MRR & H@1 & H10 & MRR & H1 & H@10 & MRR \\
    \hline
    \bf TransE & 0\% & 13.1 & 43.7 & 24.3 & 21.1 & 48.5 & 25.3 & 13.5 & 45.0 & 24.4 & 17.5 & 48.8 & 27.6 & 7.3 & 29.3 & 16.9 \\
    \bf RotatE & 0\% & 14.5 & 36.2 & 26.2 & 26.4 & 60.2 & 39.8 & 21.2 & 53.9 & 33.8 & 23.2 & 55.5 & 35.1 & 12.3 & 30.4 & 20.7 \\
    \bf KG-BERT & 0\% & 17.3 & 40.1 & 27.3 & 26.9 & 59.8 & 38.7 & 21.9 & 54.1 & 34.0 & 23.5 & 55.9 & 35.4 & 12.9 & 31.9 & 21.0\\
    \hline
    \bf KenS & 100\% & 26.4 & 66.1 & - & 32.9 & 64.8 & - & 22.3 & 60.6 & - & 25.2 & 62.6 & - & 14.4 & 39.6 & - \\
    \bf SS-AGA & 100\% & 30.8 & 58.6 & 35.3 & 34.6 & 66.9 & 42.9 & 25.5 & 61.9 & 36.6 & 27.1 & 65.5 & 38.4 & 16.3 & 41.3 & 23.1 \\
    
    \textbf{AlignKGC} w/o SI & 50\% & 55.1 & 84.2 & 65.5 & 46.7 & 74.4 & 56.6 & 44.5 & 74.0 & 54.9 & 44.0 & 71.4 & 53.4 & 28.5 & 54.9 & 37.5 \\
    
    \textbf{AlignKGC} w/ SI & 50\% & \bf 58.2 & 88.6 & \underline{69.4} & \underline{49.3} & 78.7 & 60.1 & \underline{48.4} & 79.4 & \underline{59.5} & \bf 48.0 & 76.6 & \underline{58.0} & \bf 31.7 & 59.8 & \underline{41.3} \\
    
    \textbf{JMAC} w/o SI & 50\% & {46.8} & \underline{90.3} & {62.3} & {48.1} & \underline{85.2} & \underline{61.3} & {43.8} & \underline{83.8} & {58.0} & {35.6} & \underline{76.7} & {50.3} & {24.5} & \underline{65.3} & {38.3} \\
    
    \textbf{JMAC} w/ SI & 50\% & \underline{55.2} & \bf 97.5 & \bf 71.7 & \bf 53.3 & \bf 91.4 & \bf 66.8 & \bf 49.3 & \bf 91.3 & \bf 64.5 & \underline{45.4} & \bf 88.2 & \bf 61.0 & \underline{29.5} & \bf 72.7 & \bf 44.6 \\
    \hline 
    %%\bottomrule
    \end{tabular}
    \end{adjustbox}
    \caption{MKGC results. All metrics are reported in \%. Here, H@1 and H@10 abbreviate Hits@1 and Hits@10, respectively. ``Align.'' denotes the percentage of alignment seeds used by each model when training. 
    The first three models are monolingual baselines (i.e. equivalent to the ``Align.'' rate of 0\%), while the remaining models are multilingual ones. {KenS} \cite{chen2020multilingual} and {SS-AGA} \cite{huang2022multilingual} are proposed for MKGC only, employing all alignment seeds, i.e. their ``Align.'' rate is 100\%. 
    Results of {TransE} \cite{bordes2013translating}, {RotatE} \cite{sun2019rotate}, {KG-BERT} \cite{kgbert}, {KenS}  and {SS-AGA}  are taken from \newcite{huang2022multilingual}, that are reported only with surface information (w/ SI). Results of {AlignKGC} \cite{singh2021multilingual} are taken from \newcite{singh2021multilingual}.}
    \label{tab:kgc_talbe}
\end{table*}

\subsection{Implementation details}\label{ssec:implementation}

The availabilty of surface information (\textbf{SI}) such as entity names makes the alignment problem less challenging \cite{xiang2021ontoea}. When surface information is not used by the methods (denoted as \textbf{w/o SI}), the entity and relation embeddings are randomly initialized. When surface information is used (denoted as \textbf{w/ SI}), initial entity and relation embeddings are obtained from pre-trained text embedding models \cite{singh2021multilingual,huang2022multilingual}. Therefore, we also evaluate these two problems and refer to them as \textbf{JMAC w/o SI} and \textbf{JMAC w/ SI}.  %Details about hyper-parameters such as $K$, $\gamma_c$, $\gamma_a$ and $\beta$ can be found in appendix A.

We implement our model using  Pytorch \cite{NEURIPS2019_9015}.
We iteratively train our JMAC components up to 30 epochs with two Adam optimizers \cite{kingma2014adam}. We use a grid search to choose the number of GNN hidden layers $K \in \{1, 2, 3\}$, the initial Adam learning rates $ \lambda \in \left\{1e^{-4}, 5e^{-4}, 1e^{-3} \right\}$, the controllable hyper-parameter $\beta\in \{0.1, 0.2, 0.3\}$ from \autoref{eqn:num_sample}, the margin hyper-parameters $\gamma_c$ and $\gamma_a$ $\in \{0, 5, 10\}$, and the input dimension and $\mathsf{MLP}$ hidden sizes $n \in \left\{128, 256, 512 \right\}$.  
The test set results for the two tasks are reported for the model checkpoint which obtains the highest MRR on the validation set of the MKGC task.

\begin{table*}[!t]
    \begin{adjustbox}{width=440pt,center}
    \centering
    \setlength{\tabcolsep}{0.2em}
    \def\arraystretch{1.15}
    \begin{tabular}{l|ccc|ccc|ccc|ccc|ccc}
    \hline
    \multirow{2}{*}{\bf Variants} & 
    \multicolumn{3}{c|}{\bf Greek} &
    \multicolumn{3}{c|}{\bf Japanese} &
    \multicolumn{3}{c|}{\bf French} &
    \multicolumn{3}{c|}{\bf Spanish} &
    \multicolumn{3}{c}{\bf English} \\
    \cline{2-16} 
    & H@1 & H@10 & MRR & H@1 & H@10 & MRR & H@1 & H@10 & MRR & H@1 & H@10 & MRR& H@1 & H@10 & MRR  \\
    \hline
    \textbf{JMAC} w/ SI & \textbf{55.2} & \textbf{97.5} & \textbf{71.7} & \textbf{53.3} & \textbf{91.4} & \textbf{66.8} & \textbf{49.3} & \textbf{91.3} & \textbf{64.5} & \textbf{45.4} & \textbf{88.2} & \textbf{61.0} & \textbf{29.5} & \textbf{72.7} & \textbf{44.6} \\
    \hline
    \ \ \ \ (i) w/o RA-GNN & 53.3 & 95.3 & 70.6 & 49.1 & 90.7 & 63.8 & 46.4 & 88.7 & 61.0 & 43.6 & 85.4 & 59.3 & 27.1 & 71.1 & 42.3  \\
    \ \ \ \ (ii)  w/ 1-GNN & 51.1 & 92.4 & 64.2 & 49.1 & 90.3 & 63.1 & 46.1 & 87.1 & 60.8 & 41.4 & 82.4 & 56.1 & 27.2 & 68.3 & 41.6 \\
    %\hline
    \ \ \ \ (iii) w/o SIR & \underline{54.8} & \underline{96.8} & \underline{71.1} & \underline{50.9} & \underline{91.1} & \underline{64.3} & \underline{48.2} & \underline{90.3} & \underline{63.6} & \underline{44.9} & \underline{86.3} & \underline{60.5} & \underline{28.4} & \underline{71.4} & \underline{43.3} \\
    \ \ \ \ (iv) w/o EnTr & 41.5 & 87.1 & 57.3 & 39.7 & 80.6 & 54.4 & 40.5 & 80.3 & 54.2 & 36.3 & 76.3 & 50.5 & 27.2 & 67.3 & 39.3 \\
    \ \ \ \ (v) w/o Align.  & 35.3 & 83.4 & 52.1 & 35.3 & 75.8 & 48.8 & 35.3 & 76.8 & 49.9 & 29.3 & 69.1 & 46.3 & 23.5 & 55.8 & 34.8 \\
    \hline
    \end{tabular}
    \end{adjustbox}
    \caption{Ablation study for the MKGC task.  \textbf{(i) w/o RA-GNN}: Without using the relation-aware message (Equation \ref{eqn:ramessage}) and relation-aware attention (Equation \ref{eqn:ra_attention}), here we replace our proposed RA-GNNs by the graph isomorphism networks  \cite{xu2018how}. \textbf{(ii)  w/ 1-GNN}: Both the Completion and Alignment components share the same relation-aware GNN encoder, also leading to ``w/o SIR''. \textbf{(iii) w/o SIR}: Without using the structural inconsistency reduction mechanism described in Section \ref{ssec:alignment}, i.e. equations \ref{eqn:aesir} and \ref{eqn:arsir} are not used. \textbf{(iv) w/o EnTr}: Without using the alignment seed enlargement and triple transferring mechanism described in Section \ref{ssec:alignment}.  \textbf{(v) w/o Align.}: Model variant containing only the Completion component without the Alignment one.}
    \label{tab:ablation_kgc_official}
\end{table*}

\begin{figure*}[!t]
    \centering
      \begin{subfigure}{0.49\linewidth}
        \centering
        \includegraphics[scale=0.49]{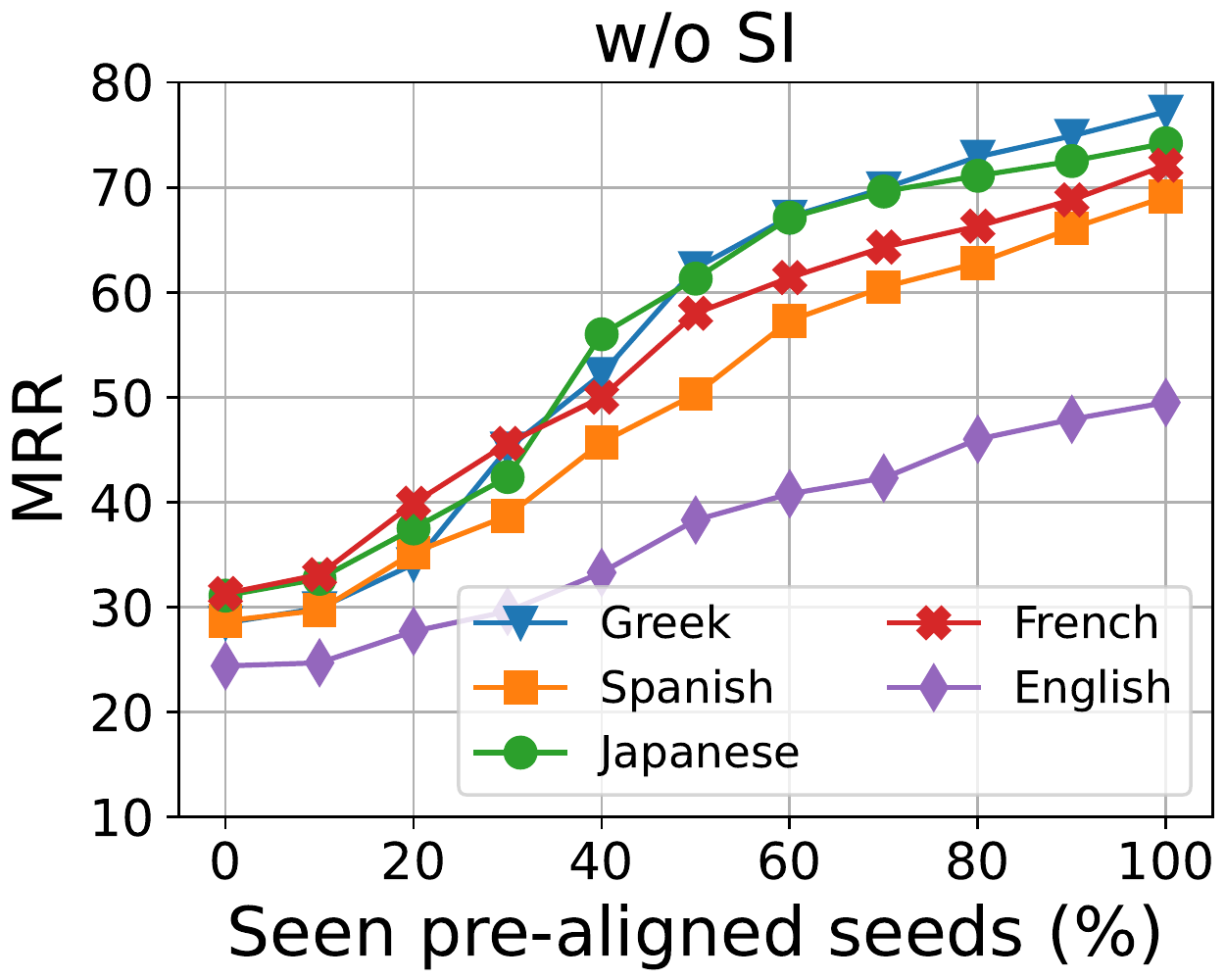}
        %\quad
        \end{subfigure}
      \begin{subfigure}{0.49\linewidth}
        \centering
        \includegraphics[scale=0.49]{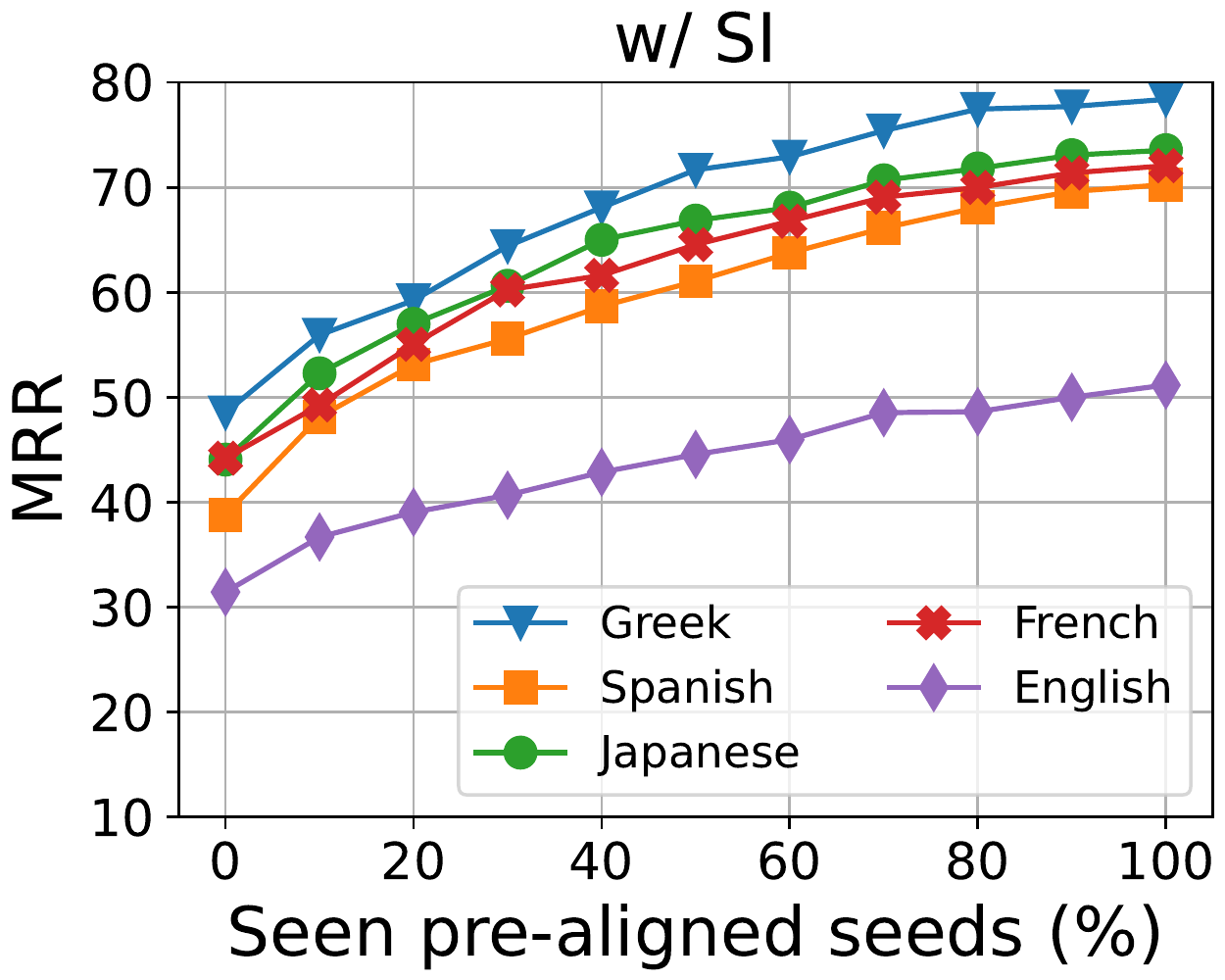}
        % \caption{MRR}
        % \label{fig:spanish_alignpc}
        \end{subfigure}
        %\quad
    \caption{MKGC MRR results w.r.t different sampling percentages of alignment seed pairs.}
    \label{fig:labeleffort}
 \end{figure*}

\section{Main Results}

\subsection{MKGC results}

\paragraph{Comparison results} \autoref{tab:kgc_talbe} lists MKGC results for JMAC and other strong baselines on the DBP-5L test sets. 
Overall, the multilingual models perform better than the monolingual ones.
It is not surprising that  JMAC and AlignKGC without surface information (w/o SI) obtain lower numbers than their counterparts using SI (w/ SI). For example, with SI information, JMAC achieves an average improvement of about 8\% points for Hits@$10$. Note that, our ``{JMAC} w/o SI'' still produces higher Hits@10 results than ``AlignKGC w/ SI'' on all KGs, and in general, performs better than ``AlignKGC w/o SI''.  Compared to {SS-AGA} that uses both SI and all available alignment seeds for training, our ``{JMAC} w/o SI'' Hits@10 results are about 30\% better for Greek, 20\% better for Japanese, French and English, and 10\% better for Spanish. We find that our ``{JMAC} w/ SI''  obtains the highest results across all KGs on all evaluation metrics, producing new state-of-the-art performances (except the second-highest Hits@1 scores on Greek, Spanish and English where ``AlignKGC w/ SI'' produces the highest Hits@1 scores).

\paragraph{Ablation study} \autoref{tab:ablation_kgc_official} presents ablation results. Removing or replacing any component or mechanism decreases the model's performance. 
The largest decrease is observed without the use of the Alignment component (w/o Align.) where the MRR scores drop about 15\% points on average. The model also performs substantially poorer when it is not using the alignment seed enlargement and triple transferring mechanism (w/o EnTr). Here, the model's Hits@$10$ scores decrease about 10\% in Greek, Japanese, French and Spanish. The performance drops incurred by using the same relation-aware GNN encoder (w/ 1-GNN) or not using relation-aware messages and attention (w/o RA-GNN) also demonstrate the importance of the two-component architecture design and the relation-ware message passing scheme. Although the structural inconsistency reduction mechanism aims to improve the alignment performance, we find that without it (w/o SIR), the MKGC performance also declines. This shows that the improvement in the MKGA task can lead to direct improvements in the MKGC task.

\paragraph{Impact of alignment seeds} To have a better insight into how much MKGA can aid MKGC,  we evaluate the MKGC task when using alignment seeds with a sampling percentage ranging from 0\% to 100\% (i.e. using all pre-aligned entity pairs). \autoref{fig:labeleffort} shows the  MRR scores obtained for this experiment.
Overall, our JMAC is improved when using more alignment seeds. 
The surface information (SI), e.g. entity names,
provides informative clues about the similarity between entities across KGs (e.g. two
entities with similar names are more likely to be an alignment pair). It is therefore not
surprising that ``{JMAC} w/ SI'' performs better than ``{JMAC} w/o SI'' in all scenarios, especially when the used sampling percentage of alignment seeds is small (i.e. SI becomes more valuable). In addition, the completion performance gradually closer as the sampling percentage  approaches 100\%. The reason is possibly that when the set of used alignment seeds is large enough, there is not much for surface information  to contribute to the alignment performance, which aids the completion performance.

%{\small
\begin{table}[!t]
    %\begin{adjustbox}{width=190pt,center}
    \centering
    \setlength{\tabcolsep}{0.4em}
    \def\arraystretch{1.1}
    \begin{tabular}{l|c|c|c}
    \hline
    \multirow{2}{*}{\bf Method} & 
    \multicolumn{3}{c}{\bf Overall} \\
    \cline{2-4}
    & Hits@1 & Hits@10 & MRR \\
    \hline
    \multicolumn{4}{c}{\textbf{w/o SI}} \\
    \hline
    \textbf{AlignKGC} & 50.2 & 65.4 & - \\
    \bf MTransE &  28.2 & 44.0 & 36.2 \\
    \bf AliNet & \underline{61.3} & \underline{73.2} & \underline{60.8} \\
    \textbf{JMAC} & \bf {63.8} & \bf 75.8 & \bf 70.3 \\
    %\hline
    \hline
    \multicolumn{4}{c}{\textbf{w/ SI}} \\
    \hline
    \textbf{AlignKGC}  & 84.8 & 91.9 & - \\
    \bf SS-AGA & 34.1 & 40.1 & 37.4 \\
    \bf PSR & 77.2 & 88.4 & 81.2 \\
    \bf RDGCN & \underline{89.3} & \underline{94.9} & \underline{91.9} \\
    \textbf{JMAC}  & \bf 93.4 & \bf 97.5 & \bf 95.1 \\
    \hline 
    %\bottomrule
    \end{tabular}
   
   % \end{adjustbox}
    \caption{MKGA results. Results for AlignKGC are taken from \newcite{singh2021multilingual}. We report our results for other baselines including {MTransE} \cite{MTransE}, {AliNet} \cite{sun2020knowledge}, {SS-AGA}, PSR \cite{Mao2021PSR} and  {RDGCN} \cite{wu2019relation}, employing their publicly released implementations. See the Appendix for the training protocols of these baselines. SS-AGA is originally proposed and evaluated for MKGC only. However, it also computes and defines an alignment matrix, thus we can also evaluate SS-AGA for the MKGA task using this matrix. 
    }
    \label{tab:end2end_kga2}
\end{table}
%}

\begin{table}[!t]
    \begin{adjustbox}{width=210pt,center}
    \centering
    \setlength{\tabcolsep}{0.4em}
    \def\arraystretch{1.1}
    \begin{tabular}{l|c|c|c}
    \hline
    \multirow{2}{*}{\bf Variants} & 
    \multicolumn{3}{c}{\bf Overall} \\
    \cline{2-4}
    & Hits@1 & Hits@10 & MRR \\
    \hline
    % \multicolumn{4}{c}{\textbf{Without SI}} \\
    \textbf{JMAC} w/ SI & \bf 93.4 & \bf 97.5 & \bf 95.1 \\
    \hline
    \ \ \ \ (i) w/o RA-GNN & 89.3 & 92.2 & 90.9 \\
    \ \ \ \ (ii) w/ 1-GNN & 62.4 & 65.4 & 64.0\\
    \ \ \ \ (iii) w/o SIR & 91.3 & 94.3 & 92.6 \\
    \ \ \ \ (iv) w/o EnTr & \underline{92.9} & \underline{95.6} & \underline{94.5} \\
    \ \ \ \ (v) w/o Comple. & 91.3 & 94.3 & 92.6 \\
    \hline
    %\bottomrule
    \end{tabular}
    \end{adjustbox}
    \caption{Ablation study for the MKGA task. \textbf{(v) w/o Comple.}: Model variant containing only the Alignment component without the Completion one. }
    \label{tab:ablation_kga_new}
\end{table}

\subsection{MKGA results}\label{ssec:mkgaresults}

\paragraph{Comparison results} \autoref{tab:end2end_kga2} presents the overall results of different models on the MKGA task. We refer the reader to the Appendix for results on each language pair. 
Overall, JMAC performs the best in both the ``w/ SI'' and ``w/o SI'' categories. Although SS-AGA performs well in the MKGC task, it performs much worse on the alignment task. Specifically, although using SI, it produces the lowest Hits@$10$. {AlignKGC}, on the other hand, achieves the third-best results in both categories. However, it is still outperformed by pure KGA models such as {AliNet} (obtaining 11.1\% points higher Hits@1 than AlignKGC in the ``w/o SI'' category) and {RDGCN} (obtaining 4.5\% points higher Hits@1 than AlignKGC in the ``w/ SI'' category). %We put the detailed comparisons for every language pairs in the Appendix. 

\paragraph{Ablation study} \autoref{tab:ablation_kga_new} shows the ablation results on the MKGA task. 
The model performance drops by about 5.3\% points in Hits@$10$ without the relation-aware messages and attention (w/o RA-GNN), confirming that the relation-aware mechanism is a crucial part of our model. 
Using only one GNN encoder (w/ 1-GNN) for both Completion and Alignment components  performs worse. This indicates that combining the objective functions and using the same feature for multiple tasks might not be optimal. 
The EnTr mechanism improves the model Hits@$10$ by about 2\% points (w/o EnTr 95.6\%  $\rightarrow$ 97.5\%). 
In the absence of SIR (w/o SIR), the alignment performance suffers a drop in each evaluation metric (2.1\% points for Hits@$1$, 3.2\% points for Hits@$10$, and 2.5\% points for MRR). This indicates that the structural inconsistency reduction improves alignment performance. 
As the Completion component only aids the Alignment component through the SIR mechanism, the model variant without Completion component (w/o Comple.)  has the same alignment performance as the model variant ``w/o SIR''.

\section{Conclusion}

We proposed JMAC, a method for joint multilingual knowledge graph completion and alignment. JMAC consists of a Completion and Alignment component and uses a new class of relation-aware GNNs for learning entity and relation embeddings suitable for both the completion and alignment tasks. We also propose a structural inconsistency reduction mechanism that fuses entity and relation embeddings from the Completion component with those of the Alignment component. We also introduce another mechanism to enlarge alignment seeds and transfer triples among KGs during training. Extensive experiments using the benchmark DBP-5L \cite{chen2020multilingual} show that JMAC performs better than previous strong baselines, producing state-of-the-art results. We publicly release the implementation of our JMAC at \url{https://github.com/vinhsuhi/JMAC}.

\section*{Limitations}

Our experimental results demonstrate that our JMAC model effectively solves the structural inconsistency problem. However, our model works based on the assumption that the surface information (SI), i.e. entity name, of an entity across different KG languages is the same or similar. In fact, some entities might have very different SI compared to their versions across different KGs, due to incorrect annotations during KGs construction or different description caused by the language barrier. We refer this issue to as SI inconsistency. When the level of SI inconsistency is high, using SI information might have reversed impacts on the model performance. A robust model should be able to decide how much it can rely on the SI. Our model has no mechanism to perform that at the moment.

\bibliography{ref}

\begin{thebibliography}{40}
\expandafter\ifx\csname natexlab\endcsname\relax\def\natexlab#1{#1}\fi

\bibitem[{Bordes et~al.(2013)Bordes, Usunier, Garcia-Dur\'{a}n, Weston, and
  Yakhnenko}]{bordes2013translating}
Antoine Bordes, Nicolas Usunier, Alberto Garcia-Dur\'{a}n, Jason Weston, and
  Oksana Yakhnenko. 2013.
\newblock {Translating Embeddings for Modeling Multi-Relational Data}.
\newblock In \emph{NIPS}.

\bibitem[{Chen et~al.(2017)Chen, Tian, Yang, and Zaniolo}]{MTransE}
Muhao Chen, Yingtao Tian, Mohan Yang, and Carlo Zaniolo. 2017.
\newblock {Multilingual Knowledge Graph Embeddings for Cross-Lingual Knowledge
  Alignment}.
\newblock In \emph{IJCAI}.

\bibitem[{Chen et~al.(2020)Chen, Chen, Fan, Uppunda, Sun, and
  Zaniolo}]{chen2020multilingual}
Xuelu Chen, Muhao Chen, Changjun Fan, Ankith Uppunda, Yizhou Sun, and Carlo
  Zaniolo. 2020.
\newblock {Multilingual Knowledge Graph Completion via Ensemble Knowledge
  Transfer}.
\newblock In \emph{Findings of EMNLP}.

\bibitem[{Dettmers et~al.(2018)Dettmers, Minervini, Stenetorp, and
  Riedel}]{dettmers2018convolutional}
Tim Dettmers, Pasquale Minervini, Pontus Stenetorp, and Sebastian Riedel. 2018.
\newblock {Convolutional 2D Knowledge Graph Embeddings}.
\newblock In \emph{AAAI}.

\bibitem[{Guo et~al.(2018)Guo, Zhang, Ge, Hu, and Qu}]{guo2018dskg}
Lingbing Guo, Qingheng Zhang, Weiyi Ge, Wei Hu, and Yuzhong Qu. 2018.
\newblock {DSKG: A deep sequential model for knowledge graph completion}.
\newblock In \emph{CCKS}.

\bibitem[{Huang et~al.(2022)Huang, Li, Jiang, Cao, Lu, Yin, Subbian, Sun, and
  Wang}]{huang2022multilingual}
Zijie Huang, Zheng Li, Haoming Jiang, Tianyu Cao, Hanqing Lu, Bing Yin, Karthik
  Subbian, Yizhou Sun, and Wei Wang. 2022.
\newblock {Multilingual Knowledge Graph Completion with Self-Supervised
  Adaptive Graph Alignment}.
\newblock In \emph{ACL}.

\bibitem[{Ji et~al.(2021)Ji, Pan, Cambria, Marttinen, and Yu}]{Ji_2021}
Shaoxiong Ji, Shirui Pan, Erik Cambria, Pekka Marttinen, and Philip~S. Yu.
  2021.
\newblock {A Survey on Knowledge Graphs: Representation, Acquisition, and
  Applications}.
\newblock \emph{TNNLS}.

\bibitem[{Jung et~al.(2020)Jung, Son, and Lyu}]{jung2020attnio}
Jaehun Jung, Bokyung Son, and Sungwon Lyu. 2020.
\newblock {AttnIO: Knowledge Graph Exploration with In-and-Out Attention Flow
  for Knowledge-Grounded Dialogue}.
\newblock In \emph{EMNLP}.

\bibitem[{Kingma and Ba(2014)}]{kingma2014adam}
Diederik Kingma and Jimmy Ba. 2014.
\newblock {Adam: A Method for Stochastic Optimization}.
\newblock \emph{arXiv preprint, arXiv:1412.6980}.

\bibitem[{Kollias et~al.(2011)Kollias, Mohammadi, and Grama}]{greedymatch}
Giorgos Kollias, Shahin Mohammadi, and Ananth Grama. 2011.
\newblock {Network Similarity Decomposition (NSD): A Fast and Scalable Approach
  to Network Alignment}.
\newblock \emph{TKDE}.

\bibitem[{Lehmann et~al.(2015)Lehmann, Isele, Jakob, Jentzsch, Kontokostas,
  Mendes, Hellmann, Morsey, van Kleef, Auer, and Bizer}]{LehmannIJJKMHMK15}
Jens Lehmann, Robert Isele, Max Jakob, Anja Jentzsch, Dimitris Kontokostas,
  Pablo~N. Mendes, Sebastian Hellmann, Mohamed Morsey, Patrick van Kleef,
  S{\"{o}}ren Auer, and Christian Bizer. 2015.
\newblock {DBpedia - A Large-scale, Multilingual Knowledge Base Extracted from
  Wikipedia}.
\newblock \emph{SWJ}.

\bibitem[{Lin et~al.(2015)Lin, Liu, Sun, Liu, and Zhu}]{lin2015learning}
Yankai Lin, Zhiyuan Liu, Maosong Sun, Yang Liu, and Xuan Zhu. 2015.
\newblock {Learning Entity and Relation Embeddings for Knowledge Graph
  Completion}.
\newblock In \emph{AAAI}.

\bibitem[{Liu et~al.(2017)Liu, Wu, and Yang}]{liu2017analogical}
Hanxiao Liu, Yuexin Wu, and Yiming Yang. 2017.
\newblock {Analogical Inference for Multi-Relational Embeddings}.
\newblock In \emph{ICML}.

\bibitem[{Mao et~al.(2021)Mao, Wang, Wu, and Lan}]{Mao2021PSR}
Xin Mao, Wenting Wang, Yuanbin Wu, and Man Lan. 2021.
\newblock {Are Negative Samples Necessary in Entity Alignment? An Approach with
  High Performance, Scalability and Robustness}.
\newblock In \emph{CIKM}.

\bibitem[{Navigli and Ponzetto(2010)}]{navigli-ponzetto-2010-babelnet}
Roberto Navigli and Simone~Paolo Ponzetto. 2010.
\newblock {{B}abel{N}et: Building a Very Large Multilingual Semantic Network}.
\newblock In \emph{ACL}.

\bibitem[{Nguyen et~al.(2018)Nguyen, Nguyen, Nguyen, and
  Phung}]{nguyen2017novel}
Dai~Quoc Nguyen, Tu~Dinh Nguyen, Dat~Quoc Nguyen, and Dinh Phung. 2018.
\newblock {A Novel Embedding Model for Knowledge Base Completion Based on
  Convolutional Neural Network}.
\newblock In \emph{NAACL}.

\bibitem[{Nguyen et~al.(2022)Nguyen, Tong, Phung, and Nguyen}]{nguyen2021node}
Dai~Quoc Nguyen, Vinh Tong, Dinh Phung, and Dat~Quoc Nguyen. 2022.
\newblock {Node Co-occurrence based Graph Neural Networks for Knowledge Graph
  Link Prediction}.
\newblock In \emph{WSDM}.

\bibitem[{Nguyen(2020)}]{Nguyen2020KGC}
Dat~Quoc Nguyen. 2020.
\newblock A survey of embedding models of entities and relationships for
  knowledge graph completion.
\newblock In \emph{TextGraphs}.

\bibitem[{Paszke et~al.(2019)Paszke, Gross, Massa et~al.}]{NEURIPS2019_9015}
Adam Paszke, Sam Gross, Francisco Massa, et~al. 2019.
\newblock Pytorch: An imperative style, high-performance deep learning library.
\newblock In \emph{NeurIPS}.

\bibitem[{Rotmensch et~al.(2017)Rotmensch, Halpern, Tlimat, Horng, and
  Sontag}]{rotmensch2017learning}
Maya Rotmensch, Yoni Halpern, Abdulhakim Tlimat, Steven Horng, and David
  Sontag. 2017.
\newblock {Learning a Health Knowledge Graph from Electronic Medical Records}.
\newblock \emph{Scientific Reports}.

\bibitem[{Schlichtkrull et~al.(2018)Schlichtkrull, Kipf, Bloem, van~den Berg,
  Titov, and Welling}]{SchlichtkrullKB17}
Michael~Sejr Schlichtkrull, Thomas~N. Kipf, Peter Bloem, Rianne van~den Berg,
  Ivan Titov, and Max Welling. 2018.
\newblock {Modeling Relational Data with Graph Convolutional Networks}.
\newblock In \emph{ESWC}.

\bibitem[{Shang et~al.(2019)Shang, Tang, Huang, Bi, He, and
  Zhou}]{shang2019end}
Chao Shang, Yun Tang, Jing Huang, Jinbo Bi, Xiaodong He, and Bowen Zhou. 2019.
\newblock {End-to-end Structure-Aware Convolutional Networks for Knowledge Base
  Completion}.
\newblock In \emph{AAAI}.

\bibitem[{Singh et~al.(2021)Singh, Jain, Mausam, and
  Chakrabarti}]{singh2021multilingual}
Harkanwar Singh, Prachi Jain, Mausam, and Soumen Chakrabarti. 2021.
\newblock {Multilingual Knowledge Graph Completion with Joint Relation and
  Entity Alignment}.
\newblock In \emph{AKBC}.

\bibitem[{Suchanek et~al.(2007)Suchanek, Kasneci, and Weikum}]{Suchanek:2007}
Fabian~M. Suchanek, Gjergji Kasneci, and Gerhard Weikum. 2007.
\newblock {YAGO: A Core of Semantic Knowledge}.
\newblock In \emph{WWW}.

\bibitem[{Sun et~al.(2020{\natexlab{a}})Sun, Zhou, and Zong}]{sun2020dual}
Jian Sun, Yu~Zhou, and Chengqing Zong. 2020{\natexlab{a}}.
\newblock {Dual Attention Network for Cross-lingual Entity Alignment}.
\newblock In \emph{COLING}.

\bibitem[{Sun et~al.(2020{\natexlab{b}})Sun, Wang, Hu, Chen, Dai, Zhang, and
  Qu}]{sun2020knowledge}
Zequn Sun, Chengming Wang, Wei Hu, Muhao Chen, Jian Dai, Wei Zhang, and Yuzhong
  Qu. 2020{\natexlab{b}}.
\newblock {Knowledge Graph Alignment Network with Gated Multi-hop Neighborhood
  Aggregation}.
\newblock In \emph{AAAI}.

\bibitem[{Sun et~al.(2020{\natexlab{c}})Sun, Zhang, Hu, Wang, Chen, Akrami, and
  Li}]{sun2020benchmarking}
Zequn Sun, Qingheng Zhang, Wei Hu, Chengming Wang, Muhao Chen, Farahnaz Akrami,
  and Chengkai Li. 2020{\natexlab{c}}.
\newblock {A Benchmarking Study of Embedding-Based Entity Alignment for
  Knowledge Graphs}.
\newblock In \emph{VLDB}.

\bibitem[{Sun et~al.(2019)Sun, Deng, Nie, and Tang}]{sun2019rotate}
Zhiqing Sun, Zhi-Hong Deng, Jian-Yun Nie, and Jian Tang. 2019.
\newblock {RotatE: Knowledge Graph Embedding by Relational Rotation in Complex
  Space}.
\newblock In \emph{ICLR}.

\bibitem[{Tong et~al.(2021)Tong, Nguyen, Phung, and Nguyen}]{tong2021two}
Vinh Tong, Dai~Quoc Nguyen, Dinh Phung, and Dat~Quoc Nguyen. 2021.
\newblock {Two-view Graph Neural Networks for Knowledge Graph Completion}.
\newblock \emph{arXiv preprint}, arXiv:2112.09231.

\bibitem[{Trouillon et~al.(2016)Trouillon, Welbl, Riedel, Gaussier, and
  Bouchard}]{pmlr-v48-trouillon16}
Théo Trouillon, Johannes Welbl, Sebastian Riedel, Eric Gaussier, and Guillaume
  Bouchard. 2016.
\newblock Complex embeddings for simple link prediction.
\newblock In \emph{ICML}.

\bibitem[{Vashishth et~al.(2020)Vashishth, Sanyal, Nitin, and
  Talukdar}]{vashishth2019composition}
Shikhar Vashishth, Soumya Sanyal, Vikram Nitin, and Partha Talukdar. 2020.
\newblock {Composition-based Multi-Relational Graph Convolutional Networks}.
\newblock In \emph{ICLR}.

\bibitem[{Wang et~al.(2014)Wang, Zhang, Feng, and Chen}]{wang2014knowledge}
Zhen Wang, Jianwen Zhang, Jianlin Feng, and Zheng Chen. 2014.
\newblock {Knowledge Graph Embedding by Translating on Hyperplanes}.
\newblock In \emph{AAAI}.

\bibitem[{Wang et~al.(2018)Wang, Lv, Lan, and Zhang}]{wang2018cross}
Zhichun Wang, Qingsong Lv, Xiaohan Lan, and Yu~Zhang. 2018.
\newblock {Cross-lingual Knowledge Graph Alignment via Graph Convolutional
  Networks}.
\newblock In \emph{EMNLP}.

\bibitem[{Wu et~al.(2019)Wu, Liu, Feng, Wang, Yan, and Zhao}]{wu2019relation}
Yuting Wu, Xiao Liu, Yansong Feng, Zheng Wang, Rui Yan, and Dongyan Zhao. 2019.
\newblock {Relation-Aware Entity Alignment for Heterogeneous Knowledge Graphs}.
\newblock In \emph{IJCAI}.

\bibitem[{Xiang et~al.(2021)Xiang, Zhang, Chen, Chen, Lin, and
  Zheng}]{xiang2021ontoea}
Yuejia Xiang, Ziheng Zhang, Jiaoyan Chen, Xi~Chen, Zhenxi Lin, and Yefeng
  Zheng. 2021.
\newblock {OntoEA: Ontology-guided Entity Alignment via Joint Knowledge Graph
  Embedding}.
\newblock In \emph{Findings of ACL}.

\bibitem[{Xu et~al.(2019)Xu, Hu, Leskovec, and Jegelka}]{xu2018how}
Keyulu Xu, Weihua Hu, Jure Leskovec, and Stefanie Jegelka. 2019.
\newblock {How Powerful are Graph Neural Networks?}
\newblock In \emph{ICLR}.

\bibitem[{Xu et~al.(2018)Xu, Li, Tian, Sonobe, Kawarabayashi, and
  Jegelka}]{jumping_knowledge}
Keyulu Xu, Chengtao Li, Yonglong Tian, Tomohiro Sonobe, Ken{-}ichi
  Kawarabayashi, and Stefanie Jegelka. 2018.
\newblock {Representation Learning on Graphs with Jumping Knowledge Networks}.
\newblock In \emph{ICML}.

\bibitem[{Yang et~al.(2015)Yang, Yih, He, Gao, and Deng}]{yang2014embedding}
Bishan Yang, Wen-tau Yih, Xiaodong He, Jianfeng Gao, and Li~Deng. 2015.
\newblock {Embedding Entities and Relations for Learning and Inference in
  Knowledge Bases}.
\newblock In \emph{ICLR}.

\bibitem[{Yao et~al.(2019)Yao, Mao, and Luo}]{kgbert}
Liang Yao, Chengsheng Mao, and Yuan Luo. 2019.
\newblock {KG-BERT:} {BERT} for knowledge graph completion.
\newblock \emph{arXiv preprint}, arXiv:1909.03193.

\bibitem[{Zhao et~al.(2020)Zhao, Xiang, Zhu, Zhang, Zhou, and
  Zong}]{zhao2020knowledge}
Yang Zhao, Lu~Xiang, Junnan Zhu, Jiajun Zhang, Yu~Zhou, and Chengqing Zong.
  2020.
\newblock {Knowledge Graph Enhanced Neural Machine Translation via Multi-task
  Learning on Sub-entity Granularity}.
\newblock In \emph{COLING}.

\end{thebibliography}
\bibliographystyle{acl_natbib}

\begin{table*}[!t]
    \begin{adjustbox}{width=440pt,center}
    \centering
    \setlength{\tabcolsep}{0.3em}
    \begin{tabular}{l|c|c|c|c|c|c|c|c|c|c|c}
    \hline
    {\bf Method} & 
    \bf EL - EN & \bf EL - ES & \bf EL - FR & \bf EL - JA & \bf EN - FR & \bf ES - EN & \bf ES - FR & \bf JA - EN & \bf JA - ES & \bf JA - FR & \bf Overall \\
    \cline{2-12}
    % \hline
    \hline
    \multicolumn{12}{c}{\textbf{w/o SI}} \\
    \hline
        \textbf{AlignKGC} & - & - & - & - & - & - & - & - & - & - & 50.2 \\
\bf MTransE & 24.2 & 34.9 & 30.0 & 38.6 & 20.6 & 23.6 & 31.6 & 19.2 & 26.7 & 41.6 & 28.2 \\
    \bf AliNet & \underline{51.5} & \textbf{73.5} & \underline{53.7} & \underline{69.1} & \underline{51.7} & \underline{67.0} & \underline{70.1} & \underline{48.7} & \underline{56.2} & \textbf{75.1} & \underline{61.3} \\
    \textbf{JMAC} & \textbf{61.1} & \underline{72.8} & \textbf{64.3} & \textbf{69.3} & \textbf{59.0} & \textbf{67.2} & \textbf{72.9} & \textbf{54.2} & \textbf{59.6} & \underline{64.7} & \textbf{63.8} \\
    \hline
    \multicolumn{12}{c}{\textbf{w/ SI}} \\
    \hline
    \textbf{AlignKGC} & - & - & - & - & - & - & - & - & - & - & 84.8 \\
    \bf SS-AGA & 16.2 & 15.9 & 15.0 & 2.7 & 70.4 & 79.8 & 76.4 & 6.8 & 5.6 & 4.8 & 34.1 \\
    \bf PSR & 77.1 & 79.5 & 74.0 & 75.2 & 76.9 & 86.3 & 86.8 & 68.1 & 66.3 & 80.1 & 77.2 \\
    \bf RDGCN & \underline{93.0} & \underline{88.6} & \underline{88.0} & \underline{84.9} & \underline{89.6} & \textbf{94.0} & \underline{88.7} & \underline{91.1} & \underline{84.0} & \underline{89.0} & \underline{89.3} \\
    \textbf{JMAC} & \bf 93.1 & \bf 92.8 & \bf 92.4 & \bf 93.8 & \bf 94.5 & \underline{93.3} & \bf 94.5 & \bf 94.7 & \bf 91.9 & \bf 93.1 & \bf 93.4 \\
    \hline 
    %\bottomrule
    \end{tabular}
    \end{adjustbox}
    \caption{MKGA Hits@1 results.}
    \label{tab:end2end_kga_detail}
\end{table*}

\begin{table*}[!t]
    \begin{adjustbox}{width=440pt,center}
    \centering
    \setlength{\tabcolsep}{0.3em}
    \begin{tabular}{l|c|c|c|c|c|c|c|c|c|c|c}
    \hline
    {\bf Method} & 
    \bf EL - EN & \bf EL - ES & \bf EL - FR & \bf EL - JA & \bf EN - FR & \bf ES - EN & \bf ES - FR & \bf JA - EN & \bf JA - ES & \bf JA - FR & \bf Overall \\
    \cline{2-12}
    % \hline
    \hline
    \multicolumn{12}{c}{\textbf{w/o SI}} \\
    \hline
    \textbf{AlignKGC} & - & - & - & - & - & - & - & - & - & - & 65.4 \\
    \bf MTransE & 43.2 & 54.1 & 45.3 & 55.4 & 33.4 & 39.4 & 48.0 & 32.9 & 42.1 & 58.8 & 44.0 \\
    \bf AliNet & \underline{66.2} & \bf 82.7 & \underline{67.4} & \bf 79.6 & \underline{65.1} & \underline{77.0} & \underline{82.4} & \underline{62.2} & \underline{68.3} & \bf 84.3 & \underline{73.2} \\
    \textbf{JMAC} & \bf 68.8 & \underline{80.6} & \bf 73.0 & \underline{79.4} & \bf 71.2 & \bf 78.7 & \bf 85.6 & \bf 67.2 & \bf 74.1 & \underline{80.3} & \bf 75.8 \\
    \hline
    \multicolumn{12}{c}{\textbf{w/ SI}} \\
    \hline
    \textbf{AlignKGC} & - & - & - & - & - & - & - & - & - & - & 91.9 \\
    \bf SS-AGA & 23.3 & 24.2 & 23.1 & 5.1 & 76.1 & 86.1 & 81.4 & 13.2 & 11.8 & 10.6 & 40.1 \\
    \bf PSR & 87.9 & 91.3 & 86.4 & 88.3 & 90.3 & 92.3 & 92.2 & 86.0 & 83.0 & 86.2 & 88.4 \\
    \bf RDGCN & \underline{97.3} & \underline{95.7} & \underline{94.9} & \underline{91.1} & \underline{95.3} & \bf {97.8} & \underline{94.6} & \underline{95.1} & \underline{91.3} & \underline{94.9} & \underline{94.9} \\
    \textbf{JMAC} & \bf 97.5 & \bf 97.8 & \bf 96.8 & \bf 97.6 & \bf 97.8 & \underline{ 97.7} & \bf 97.8 & \bf 98.2 & \bf 95.8 & \bf 97.6 & \bf 97.5 \\
    \hline 
    %\bottomrule
    \end{tabular}
    \end{adjustbox}
    \caption{MKGA Hits@10 results.}
    \label{tab:end2end_kga_hits10}
\end{table*}

\begin{table*}[!t]
    \begin{adjustbox}{width=440pt,center}
    \centering
    \setlength{\tabcolsep}{0.3em}
    \begin{tabular}{l|c|c|c|c|c|c|c|c|c|c|c}
    \hline
    {\bf Method} & 
    \bf EL - EN & \bf EL - ES & \bf EL - FR & \bf EL - JA & \bf EN - FR & \bf ES - EN & \bf ES - FR & \bf JA - EN & \bf JA - ES & \bf JA - FR & \bf Overall \\
    \cline{2-12}
    % \hline
    \hline
    \multicolumn{12}{c}{\textbf{w/o SI}} \\
    \hline
    \bf MTransE & 33.5 & 44.2 & 38.0 & 47.0 & 27.4 & 31.7 & 39.6 & 26.7 & 34.4 & 49.9 & 36.2 \\
    \bf AliNet & \underline{58.4} & \bf {77.8} & \underline{60.2} & \bf {74.1} & \underline{57.9} & \underline{71.7} & \underline{75.7} & \underline{54.9} & \underline{61.8} & \bf {79.4} & \underline{60.8} \\
%    \bf AlignKGC & - & - & - & - & - & - & - & - & - & - & - \\
    \textbf{JMAC} & \bf 63.6 & \underline{74.8} & \bf 67.0 & \underline{73.7} & \bf 65.1 & \bf 72.8 & \bf 80.4 & \bf 62.4 & \bf 69.1 & \underline{74.6} & \bf 70.3 \\
    \hline
    \multicolumn{12}{c}{\textbf{w/ SI}} \\
    \hline
    \bf SS-AGA & 20.2 & 20.5 & 19.8 & 4.6 & 73.2 & 82.6 & 78.8 & 10.6 & 9.2 & 8.3 & 37.4 \\
    \bf PSR & 81.1 & 83.8 & 78.1 & 79.9 & 81.6 & 88.6 & 88.9 & 74.3 & 72.3 & 81.9 & 81.2 \\
    \bf RDGCN & \underline{95.1} & \underline{91.8} & \underline{91.3} & \underline{87.9} & \underline{92.2} & \bf {95.7} & \underline{91.4} & \underline{93.1} & \underline{87.4} & \underline{91.7} & \underline{91.9} \\
    \textbf{JMAC} & \bf 95.3 & \bf 95.0 & \bf 94.6 & \bf 95.2 & \bf 94.6 & \bf 95.7 & \bf 96.0 & \bf 96.3 & \bf 93.3 & \bf 95.3 & \bf 95.1 \\
    \hline 
    %\bottomrule
    \end{tabular}
    \end{adjustbox}
    \caption{MKGA MRR results. \newcite{singh2021multilingual} do not report the MRR results for AlignKGC.}
    \label{tab:end2end_kga_detailMRR}
\end{table*}

\begin{table*}[!t]
    \begin{adjustbox}{width=440pt,center}
    \centering
    \setlength{\tabcolsep}{0.3em}
    \begin{tabular}{l|c|c|c|c|c|c|c|c|c|c|c}
    \hline
    \textbf{Variants} & \textbf{EL-EN} & \textbf{EL-ES} & \textbf{EL-FR} & \textbf{EL-JA} & \textbf{EN-FR} & \textbf{ES-EN} & \textbf{ES-FR} & \textbf{JA-EN} & \textbf{JA-ES} &  \textbf{JA-FR} & \textbf{Overall} \\
    %\cline{2-12} 
    \hline
    \textbf{JMAC} w/ SI  & \textbf{93.1} & \textbf{92.8} & \textbf{92.4} & \textbf{93.8} & \textbf{94.5} & \textbf{93.3} & \textbf{94.5} & \textbf{94.7} & \textbf{91.9} & \textbf{93.1} & \textbf{93.4} \\
    \hline
    \ \ \ \ (i) w/o RA-GNN & 89.0 & 89.4 & 85.1 & 89.3 & 91.4 & 89.6 & 90.7 & 90.4 & 87.3 & 88.2 & 89.3 \\
    \ \ \ \ (ii) w/ 1-GNN & 60.3 & 72.1 & 63.6 & 66.8 & 59.1 & 67.1 & 71.0 & 52.3 & 56.8 & 63.8 & 62.4 \\
    \ \ \ \ (iii) w/o SIR & 90.1 & 90.4 & 87.8 & 91.7 & 92.9 & 91.3 & 92.6 & 92.5 & 89.9 & 90.8 & 91.3 \\
    \ \ \ \ (iv) w/o EnTr & \underline{92.6} & \underline{92.1} & \underline{88.9} & \underline{93.5} & \underline{94.0} & \underline{93.1} &  \underline{94.1} & \underline{93.9} & \underline{91.2} & \underline{92.9} & \underline{92.9} \\
    \ \ \ \ (v) w/o Comple. & 90.1 & 90.4 & 87.8 & 91.7 & 92.9 & 91.3 & 92.6 & 92.5 & 89.9 & 90.8 & 91.3 \\
    \hline
    \end{tabular}
    \end{adjustbox}
    \caption{Ablation Hits@1 results for the MKGA task.}
    \label{tab:ablation_kga_old}
\end{table*}

\begin{table*}[!t]
    \begin{adjustbox}{width=440pt,center}
    \centering
    \setlength{\tabcolsep}{0.3em}
    \begin{tabular}{l|c|c|c|c|c|c|c|c|c|c|c}
    \hline
    \textbf{Variants} & \textbf{EL-EN} & \textbf{EL-ES} & \textbf{EL-FR} & \textbf{EL-JA} & \textbf{EN-FR} & \textbf{ES-EN} & \textbf{ES-FR} & \textbf{JA-EN} & \textbf{JA-ES} &  \textbf{JA-FR} & \textbf{Overall} \\
    %\cline{2-12} 
    \hline
    \textbf{JMAC}  w/ SI & \bf 97.5 & \bf 97.8 & \bf 96.8 & \bf 97.6 & \bf 97.8 & \bf 97.7 & \bf 97.8 & \bf 98.2 & \bf 95.8 & \bf 97.6 & \bf 97.5 \\
    \hline
    \ \ \ \ (i) w/o RA-GNN & 91.3 & 92.1 & 91.6 & 93.3 & 92.7 & 93.4 & 92.5 & 93.1 & 90.8 & 91.3 & 92.2 \\
    \ \ \ \ (ii) w/ 1-GNN & 63.3 & 76.4 & 65.2 & 69.3 & 64.3 & 71.1 & 73.3 & 55.2 & 57.8 & 65.3 & 65.4 \\
    \ \ \ \ (iii) w/o SIR & 94.1 & 95.3 & 93.6 & 94.1 & 95.2 & 95.0 & 94.2 & 95.1 & 92.3 & 94.4 & 94.3 \\
    \ \ \ \ (iv) w/o EnTr & \underline{95.6} & \underline{95.9} & \underline{94.9} & \underline{95.0} & \underline{95.8} & \underline{96.1} & \underline{95.8} & \underline{96.2} & \underline{94.2} & \underline{96.4} & \underline{95.6}  \\
    \ \ \ \ (v) w/o Comple. & 94.1 & 95.3 & 93.6 & 94.1 & 95.2 & 95.0 & 94.2 & 95.1 & 92.3 & 94.4 & 94.3 \\
    \hline
    \end{tabular}
    \end{adjustbox}
    \caption{Ablation Hits@10 results for the MKGA task.}
    \label{tab:ablation_kga_detailhit10}
\end{table*}

\begin{table*}[!t]
    \begin{adjustbox}{width=440pt,center}
    \centering
    \setlength{\tabcolsep}{0.3em}
    \begin{tabular}{l|c|c|c|c|c|c|c|c|c|c|c}
    \hline
    \textbf{Variants} & \textbf{EL-EN} & \textbf{EL-ES} & \textbf{EL-FR} & \textbf{EL-JA} & \textbf{EN-FR} & \textbf{ES-EN} & \textbf{ES-FR} & \textbf{JA-EN} & \textbf{JA-ES} &  \textbf{JA-FR} & \textbf{Overall} \\
    %\cline{2-12} 
    \hline
    \textbf{JMAC}  w/ SI & \bf 95.3 & \bf 95.0 & \bf 94.6 & \bf 95.2 & \bf 94.6 & \bf 95.7 & \bf 96.0 & \bf 96.3 & \bf 93.3 & \bf 95.3 & \bf 95.1 \\
    \hline
    \ \ \ \ (i) w/o RA-GNN & 90.1 & 91.1 & 90.3 & 92.4 & 91.6 & 91.3 & 91.4 & 91.7 & 89.7 & 89.3 & 90.9 \\
    \ \ \ \ (ii) w/ 1-GNN & 61.4 & 73.7 & 64.6 & 67.3 & 60.1 & 68.4 & 72.6 & 53.7 & 57.6 & 64.4 & 64.0 \\
    \ \ \ \ (iii) w/o SIR & 93.2 & 93.1 & 91.4 & 91.6 & 92.4 & 93.6 & 92.7 & 93.5 & 90.3 & 93.6 & 92.6 \\
    \ \ \ \ (iv) w/o EnTr & \underline{95.0} & \underline{94.6} & \underline{93.2} & \underline{93.5} & \underline{94.3} & \underline{95.5} & \underline{94.8} & \underline{95.3} & \underline{92.3} & \underline{95.2} & \underline{94.5} \\
    \ \ \ \ (v) w/o Comple. & 93.2 & 93.1 & 91.4 & 91.6 & 92.4 & 93.6 & 92.7 & 93.5 & 90.3 & 93.6 & 92.6 \\
    \hline
    \end{tabular}
    \end{adjustbox}
    \caption{Ablation MRR results for the MKGA task.}
    \label{tab:ablation_kga_detailMRR}
\end{table*}

\begin{table*}[!h]
\centering
%\resizebox{1.5\columnwidth}{!}{
\begin{tabular}{l|l|ccc|ccc}
\hline
\multirow{2}{*}{\bf KG Pairs} & \multirow{2}{*}{\bf  KGs} & \multicolumn{3}{c|}{\bf V1} & \multicolumn{3}{c}{\bf V2} \\
\cline{3-8} 
& &  \textbf{\#Entity}  & \textbf{\#Relation}  & \textbf{\#Triple}  &  \textbf{\#Entity} &  \textbf{\#Relation} & \textbf{\#Triple}  \\
\hline
\multirow{2}{*}{\bf  EN-FR-15K} & EN & 15,000 & 267 & 47,334 & 15,000 & 193 & 96,318 \\
& FR & 15,000 & 210 & 40,864 & 15,000 & 166 & 80,112 \\
\hline
\multirow{2}{*}{\bf  EN-DE-15K} & EN & 15,000 & 215 & 47,676 & 15,000 & 169 & 84,867 \\
& DE & 15,000 & 131 & 50,419 & 15,000 & 96 & 92,632 \\
\hline
\end{tabular}
%}
\caption{Statistics of bilingual KG pairs from the OpenEA 15K benchmark DBP1.0.}
\label{tab:openeastat}
\end{table*}

\begin{table*}[!ht]
\centering
\resizebox{2.\columnwidth}{!}{
\footnotesize
    \begin{tabular}{l|l|c c c| c c c}
    \hline
    \multirow{2}{*}{\bf  KG Pairs} & \multirow{2}{*}{\bf  Metric} & \multicolumn{3}{c|}{\bf w/o SI} & \multicolumn{3}{c}{\bf w/ SI}\\ 
    \cline{3-8}
    & & \textbf{MTransE} & \textbf{AliNet} & \textbf{JMAC} & \textbf{RDGCN} & \textbf{PSR} & \textbf{JMAC}  \\
    \hline
    \multirow{3}{*}{\textbf{EN-FR-15K V1}}
& Hits@1 & 24.7 & \underline{38.8} & \bf 58.8 & {75.4} & \underline{76.5} & \bf 90.0 \\ 
& Hits@10 & 56.3 & \underline{82.9} & \bf 83.2 & {88.1} & \underline{93.2} & \bf 98.0 \\
& MRR & 35.2 & \underline{48.5} & \bf 67.4 & {80.1} & \underline{82.3} & \bf 93.0 \\
\hline
\multirow{3}{*}{\textbf{EN-FR-15K V2}}
& Hits@1 & 24.1 & \underline{58.1} & \bf 70.9 & {84.7} & \underline{92.5} & \bf 97.1  \\ 
& Hits@10 & 24.0 & \underline{87.8} & \bf 89.0 & {93.4} & \underline{98.4} & \bf 99.6 \\
& MRR & 33.7 & \underline{69.2} & \bf 77.7 & {88.0} & \underline{94.3} & \bf 98.1 \\
\hline
\multirow{3}{*}{\textbf{EN-DE-15K V1}} 
& Hits@1 & 30.8 & \underline{61.0} & \bf 73.2 & {83.0} & \underline{88.2} & \bf 94.3 \\ 
& Hits@10 & 61.1 & \underline{83.1} & \bf 90.9 & {91.4}  & \underline{95.5} & \bf 99.0 \\
& MRR & 41.0 & \underline{68.2} & \bf 79.5 & {85.6} & \underline{91.4} & \bf 96.1 \\
\hline
\multirow{3}{*}{\textbf{EN-DE-15K V2}}
& Hits@1 & 19.4 & \underline{81.5} & \bf 89.8 & 83.4 & \underline{96.5} & \bf 97.8 \\ 
& Hits@10 & 43.2 & \underline{93.0} & \bf 97.1 & {93.6}  & \underline{99.1} & \bf 99.5 \\
& MRR & 27.4 & \underline{85.6} & \bf 92.5 & 86.1 & \underline{97.7} & \bf 98.6 \\
%    \hline
    \hline
    \end{tabular}
}
\caption{DBP1.0 test set results. We report our results for AliNet and PSR using their publicly released implementations. Results for MTransE and RDGCN are taken from \newcite{sun2020benchmarking}. Here, RDGCN is the best performing model among 12 different models experimented by \newcite{sun2020benchmarking}.
}
\label{tab:bilingual}
\end{table*}

% 0.7508571428571429, 0.9146666666666666, 0.8120971174658769

\appendix 

\section{Appendix}
\label{sec:appendix}

\subsection{Training protocols for baselines}\label{ssec:implementationbl}
For the baseline alignment models listed in Table \ref{tab:end2end_kga2}, we apply the same training protocol as detailed in Section \ref{ssec:implementation} w.r.t. the optimizer, the hidden layers, the initial learning rate values and the number of training epochs. For GNN-based alignment models {AliNet}, {SS-AGA}, PSR and {RDGCN}, we also search their number of GNN layers from $\{1, 2, 3\}$. For their other hyper-parameters, we use their implementation's default values.

\subsection{MKGA results on the DBP-5L dataset}  
Tables \ref{tab:end2end_kga_detail}--\ref{tab:ablation_kga_detailMRR} detail alignment results of experimental models as well as ablation  results of our JMAC for all language pairs in the DBP-5L dataset.

\subsection{Additional results for the KGA task} 
Note that our JMAC can perform KGA on a benchmark that is purely constructed for the KGA task. To further demonstrate the effectiveness of our JMAC, we conduct an additional KGA experiment using bilingual KG pairs from the OpenEA 15K benchmark DBP1.0 \cite{sun2020benchmarking}.  
Each KG pair consists of two versions V1 and V2 which are the sparse and dense ones, respectively. The alignment seeds are divided into 20\%, 10\%, and 70\% for training, validation and test, respectively. Statistics of the bilingual KG pairs from the OpenEA 15K benchmark DBP1.0 are presented in \autoref{tab:openeastat}.  For this entity alignment experiment, training protocols of JMAC and baselines are the same as described in Sections \ref{ssec:implementation} and \ref{ssec:implementationbl}. Here, the test set results are reported for the model checkpoint which obtains the highest MRR on the validation set. Table \ref{tab:bilingual} reports obtained alignment results on the test sets, where our JMAC performs better than the baselines in both the ``w/ SI'' and ``w/o SI'' categories, obtaining new state-of-the-art performances.

\end{document}